\documentclass[sigconf,nonacm]{aamas} 

\usepackage{balance} 

\newcommand*\oline[1]{%
  \vbox{%
    \hrule height 0.5pt
    \kern0.25ex
    \hbox{%
      \kern-0.1em
      \ifmmode#1\else\ensuremath{#1}\fi
      \kern-0.1em
    }
  }
}

\usepackage{amsmath,amsthm,amsfonts,bm,ushort}
\usepackage{xfrac}  
\usepackage{etoolbox}
\usepackage[mathscr]{euscript} 
\usepackage[T1]{fontenc}

\usepackage{enumitem}
\usepackage{xparse}

\usepackage{pifont}

\usepackage{calc}
\usepackage{color}
\usepackage[dvipsnames]{xcolor}

\usepackage{soul}
\usepackage{amsthm}
\usepackage[noabbrev, capitalise]{cleveref}  

\newcommand{\titletext}[1]{Planning and Prediction for Time-Inconsistent Agents in Contextual Choice Tasks} 
\title[\titletext{0pt}]{\titletext{0pt}}

\crefname{figure}{Figure}{Figures}

\newcommand{\info}[1]{\colorbox{green!30}{#1}}
\newcommand{\vx}{\bm{x}}
\newcommand{\vc}{\bm{c}}
\newcommand{\rvepsilon}{\bm{\epsilon}}
\newcommand{\gN}{\mathcal{N}}
\newcommand{\gU}{\mathcal{U}}
\newcommand{\rmI}{\mathbf{I}}
\newcommand{\E}{\mathbb{E}}
\newcommand{\norm}[1]{\left\lVert#1\right\rVert}

\begin{document}

\title{Generative-Model Predictive Planning for Navigation in Partially Observable Environments}


\title{Generative-Model Predictive Planning for Navigation in Partially Observable Environments}

\author{Thomas Quilter}
\authornote{These authors contributed equally to this work.}
\affiliation{
  \department{Department of Computer Science}
  \institution{University of Manchester}
  \city{Manchester}
  \country{United Kingdom}
}

\author{Yifan Zhu}
\authornotemark[1]
\affiliation{
  \department{Department of Computer Science}
  \institution{Aalto University}
  \city{Espoo}
  \country{Finland}
}

\author{Guorui Quan}
\authornotemark[1]
\affiliation{
  \department{Department of Computer Science}
  \institution{University of Manchester}
  \city{Manchester}
  \country{United Kingdom}
}

\author{Mingfei Sun}
\affiliation{
  \department{Department of Computer Science}
  \institution{University of Manchester}
  \city{Manchester}
  \country{United Kingdom}
}

\author{Samuel Kaski}
\affiliation{
  \department{Department of Computer Science}
  \institution{Aalto University}
  \city{Espoo}
  \country{Finland}
}




\begin{abstract}
Navigation in partially observable environments presents a significant challenge for autonomous agents, requiring effective decision-making with limited sensory information in unknown environments. 
Belief-based methods, particularly those using neural networks to approximate the belief space, often fail to capture the inherent multimodality of belief spaces, especially in high-dimensional cases with perceptual aliasing. 
While generative models present a compelling alternative, they typically require substantial data or expert demonstrations and lack explicit mechanisms for long-term planning.
In this paper, we introduce \emph{BeliefDiffusion}, a novel framework that combines the benefits of both generation and planning. 
\emph{BeliefDiffusion} leverages diffusion models to explicitly characterize multimodal belief distributions and utilizes Model Predictive Control (MPC) to simultaneously plan ahead.
It consists of two steps: 
(1) Imagining plausible environment configurations based on observation history and (2) Planning efficient navigation strategies across an aggregated configurations.
Through extensive experiments in synthetic map environments, we demonstrate that BeliefDiffusion significantly outperforms both model-free reinforcement learning baselines and other generative approaches in navigation success rate and path efficiency. Our results validate that explicitly incorporating multimodal belief representations into planning enables more robust navigation in partially observable settings.

\end{abstract}

\maketitle 

\section{Introduction}
We study point-goal navigation in partially observable environments where an agent must reach a specified goal location without prior knowledge of the environment structure (i.e., the map). In our setting, the agent only observes a limited local region around itself along with its relative position with goal coordinates.
Navigation of this kind presents a fundamental challenge in many domains, e.g., robotics\cite{ross2008bayesian} and autonomous driving\cite{ort2018autonomous}. 
In this setting, agents must make decisions based on limited or incomplete sensory information about their surroundings.
One traditional approach to solving this challenge is through  Partially Observable Markov Decision Processes (POMDPs), 
which assumes that the agent cannot directly observe the true state but instead needs to maintain a probability ``belief'' of all possible states based on the observations received so far -- also known as Belief MDP. 
By constructing representations of environmental states that are consistent with observations, agents can make more informed decisions under uncertainty. Since the belief state serves as a sufficient statistic of the agent's history, there is no need to retain a growing record of past actions and observations, allowing decisions to rely on a fixed-size representation. Moreover, by maximizing the expected total discounted reward over the space of belief states, agents can achieve optimal decision-making.
Techniques such as simultaneous localization and mapping (SLAM) have been used to build belief representations of environments\cite{durrant2006simultaneous}. More recent approaches leverage neural network-based world models to infer environmental states from observation histories\cite{zou2024diffbev,zhang2024imagine}.

However, these belief-based methods often simplify belief representations into a unimodal point estimation, i.e., most likely state\cite{zhang2024imagine}, sacrificing the probabilistic reasoning necessary for optimal navigation in partially observable settings. 
This limitation is especially problematic in environments with perceptual aliasing and restricted sensor ranges, where multiple plausible configurations exist. In such cases, unimodal approximations fail to capture the true multimodal nature of the belief space, leading to suboptimal decision-making\cite{gupta2024efficient}.
Furthermore, these methods struggle to cope with high-dimensional belief spaces and require explicit environmental models that are difficult to acquire in complex, unknown settings~\cite{karkus2017qmdp,hoerger2019pomdp}. 
Previous attempts to overcome these limitations include the use of generative models, which stitch navigation trajectories directly without explicitly planning\cite{ha2018world,janner2022planning}, or reinforcement learning\cite{wijmans2019dd}, which focuses on model-free and end-to-end training.  
By learning a structured latent space, generative models enhance decision-making efficiency without directly interacting with the real environment. They effectively encode complex state distributions\cite{hafner2019dream}, capturing a diverse range of possible states. This makes them particularly well-suited for large, high-dimensional environments.
Despite their advantages, these models still require expert demonstrations \cite{yu2024trajectory}, or extensive offline trajectories \cite{janner2022planning}, and often ignore the need for long-term and robust planning in navigation\cite{hong2023diffused}.

In this paper, we combine the benefits of both worlds, generation and planning, and introduce \emph{BeliefDiffusion}, a novel framework that leverages diffusion models to explicitly characterize multimodal belief distributions and utilizes Model Predictive Control (MPC) to simultaneously plan ahead.
BeliefDiffusion follows a two-step process: (1) Imagining plausible environment configurations based on observation history and (2) Planning efficient navigation strategies across an aggregated configurations.
First, our model generalizes belief estimation to distribution generation, producing possible map layouts that are consistent with past observations. 
Our key insight is to treat map layout inference as a conditional generative modeling problem, 
in order to capture the multimodal distribution of plausible environments. 
By learning to generate diverse and realistic local maps, our model constructs an explicit belief representation over possible worlds.

\begin{figure*}[t]
    \centering
    \includegraphics[width=\textwidth]{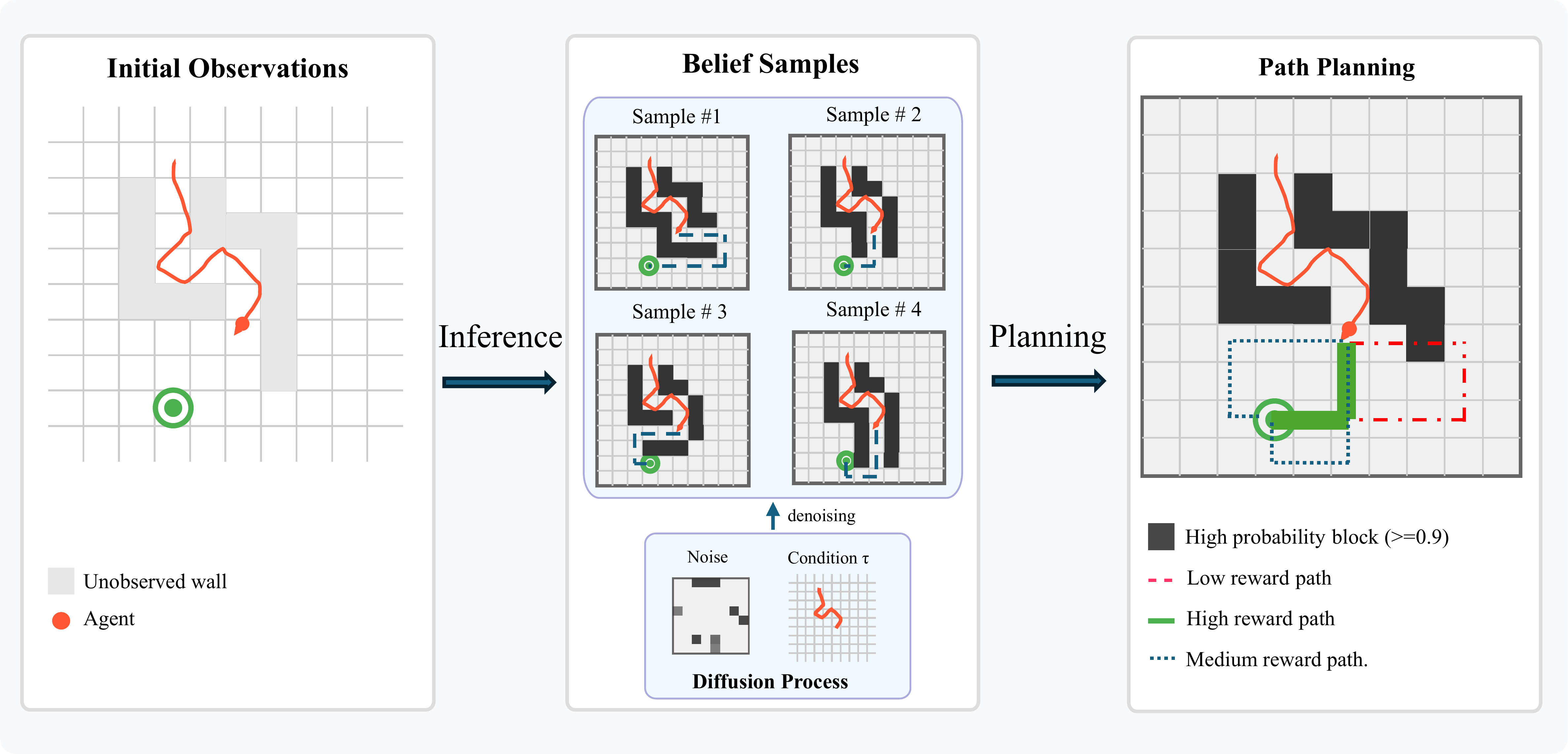}
    \caption{BeliefDiffusion explicitly characterises multimodal belief distributions using diffusion models to generate plausible environment configurations from partial observations. We then leverage Model Predictive Control, MPC, to plan robust navigation strategies across these beliefs, enabling efficient navigation in unknown environments.}
    \label{fig:teaser}
\end{figure*}

Second, we integrate this belief representation into an MPC planning framework. We iteratively minimize the cost based on a dynamic model and action proposals derived from the distribution of potential map layouts, 
and then select a locally optimal trajectory at each timestep. 
By performing path planning jointly across sampled layouts, we identify robust and efficient navigation strategies in environments where accurate long-horizon planning is impossible. \cref{fig:teaser} illustrates the key steps of BeliefDiffusion, from initial observations to belief sampling and path planning.

Through extensive experimentation in synthetic map environments, we demonstrate that BeliefDiffusion significantly outperforms existing methodologies in both navigation success rate and path efficiency. Our results show that by explicitly incorporating reasoned belief space into MPC, our framework enables more informed decision-making in complex, unknown environments without requiring the extensive training data needed by reinforcement learning approaches. The main contributions of our work include: 
\begin{itemize}
    \item A novel diffusion-based approach for explicitly modeling multimodal belief states in navigation, 
    \item A belief-space model-predictive planning framework that integrates generated map layouts for robust decision-making in navigation, 
    \item Strong empirical evidence demonstrating improved sample efficiency and performance compared to both model-free RL methods and approaches that rely on unimodal approximations.
\end{itemize}

In essence, our method first \emph{imagines} several plausible versions of the unseen parts of the world, then \emph{plans} actions that work well across those possibilities. Concretely, the diffusion model takes the agent’s partial experience, like a sketch of corridors glimpsed so far, and proposes multiple likely local maps rather than betting on a single guess. MPC then evaluates short action sequences across this small bundle of maps and picks the move that keeps progress towards the goal while avoiding routes that would fail on any reasonable alternative. This two-step “imagine then plan” loop is repeated as new observations arrive, so poor hypotheses are discarded and good ones reinforced, much like a sat-nav that continually recalculates as roads are revealed. A simple analogy: if two passages might exist ahead, we prepare for both, taking the turn that remains safe and useful whichever passage proves real. This makes the agent robust to perceptual aliasing, reduces wrong turns, and saves data, because we do not need to learn a monolithic policy for every situation, only to generate a handful of sensible maps and choose actions that hedge across them. In short, modelling uncertainty explicitly, then planning \emph{through} that uncertainty, is why BeliefDiffusion succeeds in challenging, partially observable navigation.

\section{Related Work}
\paragraph{POMDP for Partial Observability.}
Navigating in unseen environments presents significant challenges due to partial observability and unknown environmental dynamics.
Partially Observable Markov Decision Processes (POMDPs)\cite{kaelbling1998planning} provide a mathematical framework for decision-making under partial observability by maintaining belief states over possible environmental configurations. 
This enables agents to perform belief-space planning, reasoning over latent state distributions to account for perceptual aliasing and ambiguous observations. The POMDP framework has shown success in various robotic tasks involving partial observability, particularly in navigation contexts \cite{martinez2009bayesian,lauri2016planning}.
Despite their theoretical elegance, classical POMDP solvers face substantial scalability challenges. These algorithms struggle with high-dimensional belief states \cite{hauskrecht2000value} and typically require explicit transition models, which necessitate extensive data collection in complex, unknown environments \cite{karkus2017qmdp,hoerger2019pomdp}. While reinforcement learning methods such as PPO \cite{schulman2017proximal,wijmans2019dd} offer alternative approaches, they frequently suffer from data inefficiency and training instability. Recent neural approaches \cite{ramakrishnan2022poni,zhang2024imagine} attempt to model belief distributions in POMDPs but fail to address a critical challenge: capturing the multimodal nature of belief states in map environments. Instead, these methods tend to collapse to the most probable latent state or rely on a single sample from their model, rather than maintaining the full belief distribution necessary for optimal decision-making under partial observability.

\paragraph{World Models for Environmental Understanding.}
Generative models play an increasingly important role in decision-making problems, commonly referred to as World Models. These models serve multiple functions: (1) learning abstract representations of environments \citep{ha2018world}, (2) predicting future states \cite{hansen2022modem}, and finally providing contextual information for decision-making processes.
Recent advances in generative techniques, particularly diffusion models, have been applied to future state prediction \cite{janner2022planning} in complex scenarios including autonomous driving \cite{zheng2025diffusion}. These approaches effectively offer probabilistic frameworks for modeling future trajectories. However, most existing methods are constrained in scope, primarily focusing on predicting observation-wise states rather than developing holistic environmental representations.
More succinct and task-oriented representations, such as Occupancy Grids \cite{wang2024occsora} or Bird's-Eye View representations \cite{zou2024diffbev}, often prove more effective, particularly for navigation tasks. While these specialized models provide highly descriptive information, they typically lack direct integration with decision-making processes—a crucial component for effective navigation.

\section{Preliminaries}
In this section, we provide a brief introduction to Partially Observable Markov Decision Processes (POMDP), Model Predictive Control, and diffusion models with classifier-free guidance.

\subsection{Partially Observable Markov Decision Process (POMDP)}
A Partially Observable Markov Decision Process (POMDP) is formally defined as a 6-tuple $\langle \mathcal{S}, \mathcal{A}, \mathcal{O}, \mathcal{T}, \mathcal{Z}, \mathcal{R} \rangle$. Here, $\mathcal{S}$ denotes the set of states, while $\mathcal{A}$ represents the set of available actions. The agent receives observations from the set $\mathcal{O}$, which provide partial information about the underlying state. 
The system evolves according to the transition probability function $\mathcal{T}(s, a, s') = P(s_{t+1} = s' \mid s_t = s, a_t = a)$. 
The observation probability function $\mathcal{Z}(s', a, o) = P(o_{t+1} = o \mid s_{t+1} = s', a_t = a)$,  specifies the likelihood of receiving observation $o$ given that the system has transitioned to state $s'$ after taking action $a$. 
Finally, the reward function $\mathcal{R}(s, a)$ determines the immediate reward obtained upon executing action $a$ in state $s$.
A POMDP problem can be transformed into a fully observable Markov Decision Process (MDP) over belief states, often referred to as a belief MDP. 
More formally, the state space consists of all possible belief states, denoted as $b \in \Delta(\mathcal{S})$, where $\Delta(\mathcal{S})$ represents the set of all probability distributions over the state space $\mathcal{S}$. 
The action space remains the same as in the original POMDP, i.e., $\mathcal{A}$. 
The transition function, $\tau(b, a, b')$, gives the probability of transitioning to belief state $b'$ after taking action $a$ in belief state $b$. 
The reward function is defined as $\rho(b, a) = \sum_{s \in \mathcal{S}} b(s) \mathcal{R}(s, a)$, where $b(s)$ is the belief probability of state $s$ and $\mathcal{R}(s, a)$ is the reward function in the original POMDP.
When the agent takes action $a$ in belief state $b$ and receives observation $o$, the belief state is updated using Bayes' rule:
$b'(s') = \frac{P(o \mid s', a) \sum_{s \in \mathcal{S}} P(s' \mid s, a) b(s)}{P(o \mid b, a)}$
where $P(o \mid b, a)$ is given by:
$P(o \mid b, a) = \sum_{s' \in \mathcal{S}} P(o \mid s', a) \sum_{s \in \mathcal{S}} P(s' \mid s, a) b(s)$. 
Since the agent cannot directly observe the true state of the environment, it maintains a probability distribution over all possible states, known as a belief state. At time \(t\), the belief state \(b_t\) is a distribution over the state space \(\mathcal{S}\), where \(b_t(s)\) denotes the probability that the system is in state \(s \in \mathcal{S}\).

In a navigation task, an agent must move through an environment with an uncertain layout. If there is only a finite set of possible maps, denoted as \(\mathcal{M} = \{m_1, m_2, ..., m_k\}\), where each map represents a distinct arrangement of obstacles, and when the agent performs an action and moves, its observation is its current position, \(p_{obs}\). Then the state space \(\mathcal{P} \times \mathcal{M}\) is defined by the combination of the agent's position \(p\) and the actual map \(m\). Since the agent's position \(p'\) is fully observed after each action \(a\), the uncertainty about its location is eliminated, and the belief state simplifies to a probability distribution over \(\mathcal{M}\).

\subsection{Model Predictive Control}
Model Predictive Control (MPC) is a model-based method for optimal control\cite{morari1999model,schwenzer2021review}. 
The core characteristic of MPC is to optimize the reward over a receding horizon, where the objective is:  
$\hat a_{t:t+H} = \arg \max_{a_{t:t+H}} \mathbb{E}_{p_\theta(s_{t+1:t+H} | a_{t:t+H}, s_{0:t})} \left[ r_{t+1} + \dots + r_{t+H} \right]$. 
Here, a learned dynamic model \( p_\theta \) predicts future state distributions given a sequence of actions. 
\( a_{t} \) represents the action sequence over the horizon \( H \), \( s_{t+1}, \dots \) are the future states and \( r_{t+1}, \dots \) are the corresponding rewards. Once the optimal action sequence $\hat a_{t:t+H}$ is acquired, MPC use a closed-loop control method, executing the first action $a_t$ and replanning the remaining actions based on new observations.
MPC offers two key advantages:  1. Adaptability to changing environments – Since MPC iteratively updates its decisions based on new observations, it naturally reacts to uncertainties.  2. Data efficiency – Unlike reinforcement learning, MPC often requires less data because it does not rely on learning a full policy from scratch.  

Two closely related works to our approach are \textbf{Diffuser}\cite{janner2022planning} and \textbf{Diffusion Model Predictive Control (D-MPC)}\cite{zhou2024diffusion}. \textbf{Diffuser} employs a diffusion model to jointly model future state and action distributions, integrating action proposal and dynamics prediction into a single framework. In contrast, \textbf{D-MPC} separates dynamics modeling and action proposal into two diffusion models, allowing for better adaptation to environmental changes. Our method retains the advantages of separating dynamics modeling and action proposal but does so in the belief space using only a single diffusion model. This approach further enhances data efficiency while preserving adaptability.

\subsection{Denoising diffusion probabilistic models (DDPM)} 
Denoising diffusion probabilistic models (DDPM) is a generative model which have two processes: a forward diffusion process and a reverse denoising process. 
Given a sample point $\vx_0$ from an unknown distribution $p(\vx_0)$, the forward diffusion process adds a small amount of Gaussian noise to the sample $\vx_0$ in $T$ steps, producing a sequence of noisy samples $\vx_1, \vx_2,..., \vx_T$. 
The step sizes are controlled by a variance schedule $\{\beta_t\in(0, 1) \}_{t=1}^{T}$, yielding distributions
$q(\vx_t|\vx_{t-1}) = \gN(\vx_t; \sqrt{1-\beta_t}\vx_{t-1}, \beta_t\rmI)$ and
$q(\vx_{1:T}|\vx_0) = \prod_{t=1}^{T} q(\vx_t|\vx_{t-1})$. 
After defining $\alpha_t \triangleq 1 - \beta_t$ and $\bar{\alpha}_t \triangleq  \prod_{i=1}^{t} \alpha_i$, 
the diffusion kernel is given as $\vx_t = \sqrt{\bar{\alpha}_t}\vx_0 + \sqrt{1-\bar{\alpha}_t}\rvepsilon$,
where $\rvepsilon$ is the Gaussian noise.
The reverse denoising process 
reverses the diffusion process to generate data $\vx_0$ by denoising
in $T$ steps: $p(\vx_T) = \gN(\vx_T; \mathbf{0}, \rmI)$ and $p_{\theta}(\vx_{t-1}|\vx_t) = \gN(\vx_{t-1}; \mu_\theta(\vx_t, t), \sigma^2_t\rmI)$, 
where $\mu_\theta$ is a trainable network. 
After a few arithmetic operations and reparameterisation, we can write down the variational objective as:
$\min_\theta \E_{}\left[\lambda_t \norm{\rvepsilon - \rvepsilon_\theta\left( \sqrt{\bar{\alpha}_t}\vx_0 + \sqrt{1-\bar{\alpha}_t}\rvepsilon, t \right)}^2 \right]$,
where the expectation is over $\vx_0\sim q(\vx_0), t\sim \gU\{1, T\},\rvepsilon\sim\gN(\mathbf{0}, \rmI)$, and $\vx_t$ is substituted with $\vx_0$ using $\vx_t = \sqrt{\bar{\alpha}_t}\vx_0 + \sqrt{1-\bar{\alpha}_t}\rvepsilon$. 
\cite{ho2020denoising} observe that simply setting $\lambda_t$ to $1$ for all $t$ works best in practice. 

The denoising process generates a data distribution from noise without specific conditions. 
Classifier-Free Guidance (CFG) is a way to modify an unconditional denoising process by conditioning on a discrete random variable $\vc$ in order to control the generation process~\citep{ho2022classifier}. 
In practice,  CFG operates by mixing the predictions of a conditional diffusion model $p_{\theta}(\vx|\vc)$ and an unconditional diffusion model $p_{\theta}(\vx)$, 
both of which can be parameterised via a shared network architecture. 
For training the unconditional model, the condition is masked to a null token $\emptyset$. 
For sampling, the output at time step $t$ is given by
$\hat{\rvepsilon}(\vx_t, t, \vc) = (1+w) \cdot \rvepsilon_{\theta}(\vx_t, t, \vc) - w \cdot \rvepsilon_{\theta}(\vx_t, t, \emptyset)$.

\section{Method}

\subsection{Map Layout Inference with Diffusion Models}
In our first module, we aim to train a diffusion model that can infer an \( S \times S \) local map $m$ around the agent, conditioned on the agent's observation history $\tau$.


\textbf{Dataset Collection} To train the diffusion model, we collect a dataset $(m_n, \tau_n)$: $\tau_n$ is the agent's observation history $\{o_{n0},o_{n1},\dots,o_{nT}\}$, and $m_n$ is the ground truth environment map.
We discretized each map into grid cells, and each map instance in our dataset is represented as a 2D binary occupancy grid,  which is $S \times S$ in size and centered at the agent's current position. 
During the collection process, the agent navigates using a map that is intentionally distorted by noise. 
This noisy map is the ground truth map with each cell flipped with probability $p_{flip}$. 
We then use A* algorithm to plan waypoints for the agent's movement. This method is designed to collect observation history in scenarios where the agent does not have access to an accurate map during testing.

\begin{figure*}[t]
    \centering
    \includegraphics[width=\linewidth]{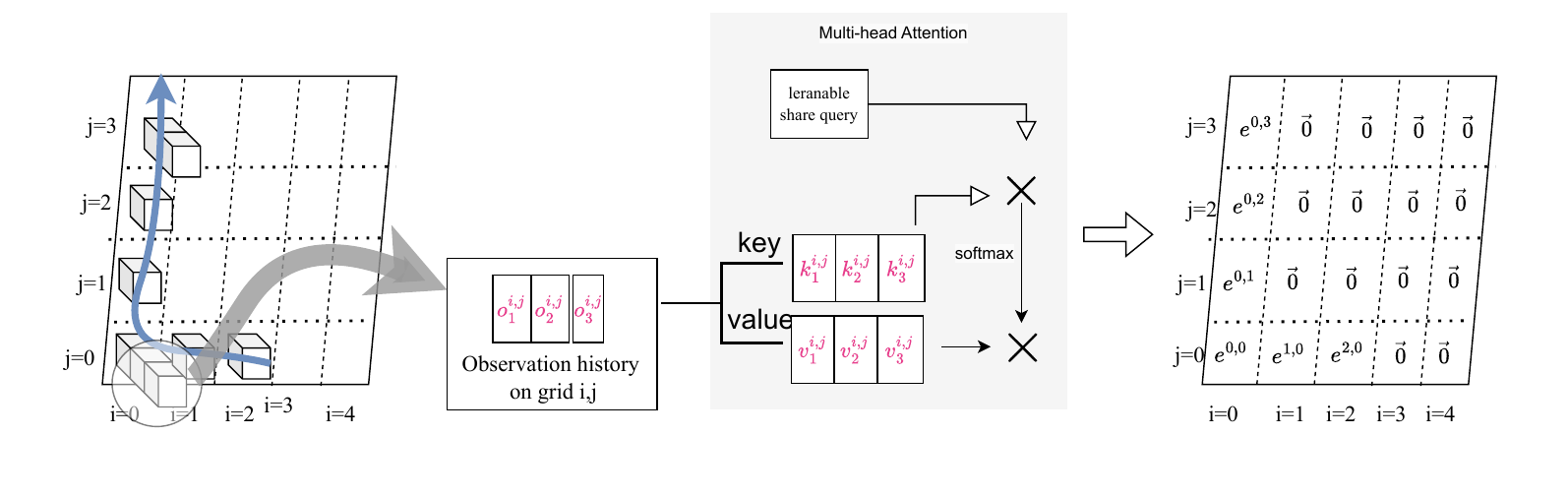}
    \caption{Map embedding relies on multi-head attention to selectively aggregate past observations to generate conditional embeddings for each grid cell in the local map.}
    \label{fig:embedding}
\end{figure*}

\textbf{Map Representation Embedding} To effectively adapting to varying history length,
we leverage multi-head attention to enable the model to focus selectively on relevant aspects of the observation history for each cell. 
This yields extra benefits when the agent remains in the same location for long periods, since simple averaging or pooling of the observation history could dilute important information. 
In specific, given a trajectory of observation history \(\tau=o_{1:T}\), we aggregate observations \(o_{t}\) based on the cell position \((i_{t}, j_{t})\) where the agent receives those observations, as shown in \cref{fig:embedding}.
Specifically, for each grid cell \((i, j)\) in the \(S \times S\) local map \(m\), we define the history sequence as the set of all past observations that occurred in that cell: \(H^{ij} = \{o_{t} \mid i_{t} = i, j_{t} = j\}\).
If a cell has not been visited, its history remains empty, i.e., \(H^{ij} = \emptyset\). 
To effectively condition the U-Net, we convert the sequence in each grid into an embedding through a multi-head attention. 
More specifically, for each grid cell \((i, j)\), we map its observation sequence \( H^{ij} \) to key \(\mathbf{k}^{i,j}_t\) and value \(\mathbf{v}^{i,j}_t\). 
A shared learnable query \(\mathbf{q}\) attends to these via multi-head attention. Unvisited cells (\( o^{i,j}_t = \emptyset \)) are assigned zero embeddings. 
The resulting embeddings \( \{\mathbf{e}_{ij}\}_{i,j \in S \times S} \) form the conditional input patch for the diffusion model. 

\textbf{Diffusion Model Training}
We utilize Denoising Diffusion Probabilistic Models (DDPM)\cite{ho2020denoising} to train our map inference model because DDPM can effectively capture multi-modality distribution, and are less susceptible to mode collapse than Generative Adversarial Networks (GANs)\cite{dhariwal2021diffusion}. 
We implement a U-Net\cite{Ronneberger2015unet} as the backbone in the diffusion model. 
Following classifier-free guidance\cite{ho2022classifier}, we randomly mask \( \{\mathbf{e}_{ij}\}_{i,j \in S \times S} \) to zero with a probability of 0.1 for the unconditional model. Instead of predicting noise, We directly predict the original map \(\vx_0 = m\), similar to DALLE~\citep{ramesh2022hierarchical}. 
Our model learns to reconstruct the original map by optimizing the following objective:  
\begin{equation}  
\mathcal{L}(\theta) = \mathbb{E}_{t, \epsilon, m, \mathbf{e}} \left[ \lVert m - \epsilon_\theta(m_t, t, \mathbf{e}) \rVert^2 \right],  
\end{equation}  
where \( t \sim \mathcal{U}\{1, 2, \ldots, N\} \) represents the diffusion timestep, \( \epsilon \sim \mathcal{N}(\mathbf{0}, \mathbf{I}) \) is the noise target, and \( m_t \) is the noisy version of the map, given by \( m_t = \sqrt{\bar{\alpha}_t} m + \sqrt{1-\bar{\alpha}_t} \rvepsilon \).  
We incorporate the conditional embedding and the 2D noise input by concatenating them and feeding the combined input into the diffusion model. For sampling from the denoised imaginary layout, we first use the conditional model to obtain the model input and then follow the denoising process outlined in \cite{ho2020denoising}. Finally, we apply a threshold of 0.5 to the samples to generate a binary matrix.

\subsection{Navigation Planning}

The trained model will generate plausible map layouts based on prior observations. 
A "plausible" map layout marks as empty for those grids traversed by the agent, and marks as blocks for those where the agent encountered obstacles, and extrapolates map structure in unobserved areas based on learned distributions, conditioned on empty and block grids. 
To enable planning, we approximate the distribution of potential map layouts, denoted as $P(m|\tau)$, by sampling a set $\mathcal{S}_N$ of $N$ map samples of size $S\times S$ from the diffusion model conditioned on the observed trajectory $\tau$. 
These samples will be used for
1) estimating the transition function in local regions. 
and 2) proposing candidate actions for planning.  
\begin{itemize}
    \item  To estimate the dynamic model, we observe that the sample distribution reveals consensus on block placements near observed regions while exhibiting diversity in unexplored areas. To model dynamics, we conservatively aggregate all generated local maps to form one map \( m_\text{agg} \) by treating a cell as a block if it is marked as a block in at least \( N \cdot \lambda_p \) of the generated samples (\( \lambda_p \) is a threshold), and as empty otherwise, even for areas outside of the generated local maps. As shown in \cref{fig:teaser}, the areas outside of the local maps are all marked as empty (in grey). 
    \item To produce action proposals, we generate a set of candidate paths, denoted as \( p_i \), by applying a planning algorithm to each individual sampled map layout. 
    Note that this step is independent of the previous dynamic prediction.
    For each sampled map, planning is conducted  with the same start position (i.e., agent's current position) and the goal position. 
\end{itemize}
All path candidates generated are then evaluated with the aggregated dynamics model, \( m_\text{agg} \): any path violating the layout in \( m_\text{agg} \) will receive zero reward. Finally, we select the path with the highest reward as the final plan.

In our planning module, the parameter \( S \) defines the size of the local map used for diffusion, effectively setting the spatial scope of our belief representation. This is analogous to the planning horizon in MPC but applied within the belief space in a spatial context. 
Importantly, \( S \) affects action proposals by constraining the spatial extent of the dynamic model during planning. 
A larger \( S \) provides a broader spatial context, allowing planning to capture more global map structures and potentially make long-term decisions,  
but comes with a price: large \( S \) yields high computational cost in inference and requires more diverse training data to learn effective representations for large map regions. 
The number of samples, \( N \), representing the number of action proposals, also introduces a trade-off. Increasing \( N \) enables a more comprehensive exploration of possible map configurations, potentially leading to more efficient paths and reducing the risk of suboptimal layouts. However, a larger \( N \)  also increases computational costs and may result in infeasible solution proposal. 

We implement the A* search algorithm for path planning by assigning a cost of 1 to moving through empty cells and penalizing inferred blocks with a cost greater than 1. Manhattan distance serves as our heuristic function. Although infinite costs for obstacles might seem ideal, we find that they can cause planning failures when inferred maps contain slight disconnections. The selected path is then converted into waypoints for the agent to follow using a greedy strategy. At each step, the agent chooses the action (move forward, turn left, or turn right) that best advances towards the next waypoint. For complex movement, more sophisticated waypoint-following or reactive control could improve performance. 

During testing, our system operates iteratively as follows: the agent collects observations, the layout inference module generates plausible map configurations, the navigation planning module plans the optimal path, and the agent executes actions to reach the next waypoints and gather new information. Once the agent reaches the first waypoint, a new round of generation and planning starts. This procedure repeats until the agent either reaches the goal or exceeds the predetermined step limit.

\section{Experiments}

\subsection{Setup of the Experiment}

\textbf{Map Dataset} We constructed a synthetic map dataset inspired by point-goal navigation in indoor environments\cite{savva2019habitat}. Each map instance in our dataset is represented as a 2D binary occupancy grid. We chose to develop our own synthetic dataset rather than using existing navigation datasets from embodied AI simulations like Habitat for two reasons: 
1. Existing datasets are often integrated with too complex simulators that are difficult to customize for our specific requirements.
2. We can have precise control over dataset quality with synthetic tasks. For example, the Gibson PointNav dataset \cite{xia2018gibson} used in the Habitat simulator \cite{savva2019habitat} contains navigation tasks with notable limitations: 23\% of validation tasks can be solved by direct move towards the goal, most tasks have relatively short routes, and some scenes contain poorly reconstructed areas with holes in the floor that disrupt navigation.

In contrast, our synthetic methodology grants us fine-grained control over map properties, enabling the generation of a diverse dataset specifically designed to isolate the navigation task itself.
For the training dataset, we randomly generated 100 maps and collected 50,000 episodes, with each episode having a maximum length of 200. The generated maps are 64×64 in size, and map-generation details are provided in the supplementary material.

\textbf{Implementation Details} Our diffusion model is based on a U-Net architecture\cite{Ronneberger2015unet} that consists of two downsampling and upsampling stages, with each resolution level containing two ResNet blocks. An attention layer is applied at an 8×8 resolution, and we use group normalization and ReLU activation throughout the model, along with a 0.1 dropout rate during training. Trajectory information is integrated into both the input and skip connections through concatenation. Our diffusion training and sampling employ a linear noise schedule and utilize the DDIM sampling algorithm\cite{song2020denoising} for efficient inference. During evaluation, we perform sampling with 20 steps. Hyperparameters are reported in the supplementary material.

\subsection{Evaluation Results}

In this section, we address the following questions:
1) Can diffusion models generate maps suitable for navigation?
2) Does the explicit belief representation improve navigation performance compared to end-to-end approaches such as LSTM-PPO?
3) Can generative models that model at the observation level achieve comparable performance?
4) How do the hyperparameters of the generative model impact planning performance?

\begin{figure}[htbp]
\centering
\includegraphics[width=0.8\linewidth]{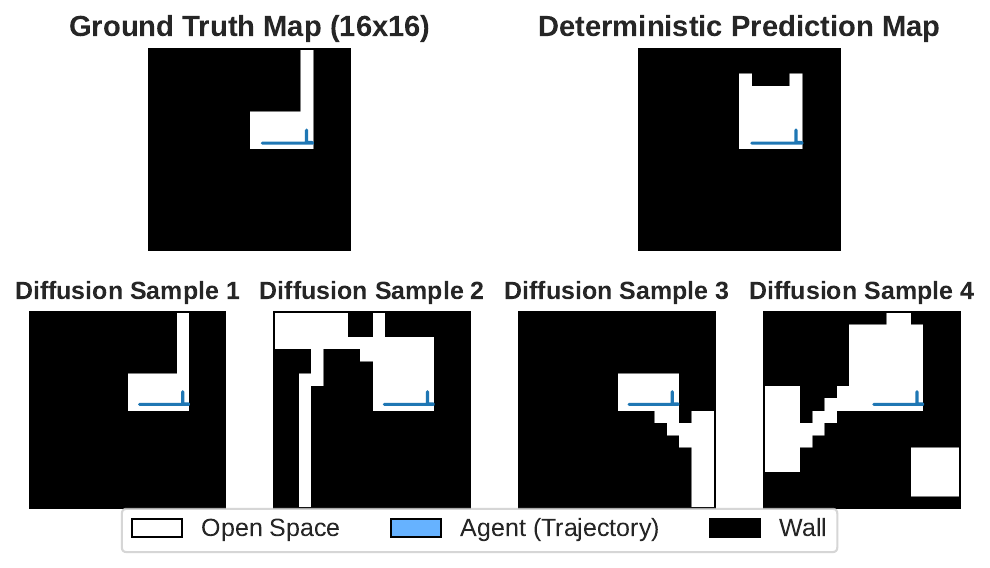}
\caption{Generated maps from diffusion model and an deterministic prediction model.}
\label{fig:map_generation}
\end{figure}

\paragraph{Diffusion Models Generate Plausible and Diverse Layouts} To assess the ability of our diffusion model to generate navigable maps, we train the diffusion model on 16×16 maps and compare it against a deterministic prediction model trained using supervised learning with an MSE objective and the same U-Net architecture. Figure~\ref{fig:map_generation} presents visualizations of map samples generated by both models. The deterministic prediction model produces maps that do not exhibit any structure in the map dataset, such as paths, failing to capture the variety of plausible layouts.

To further assess the quality of the generated maps, we conducted tests using agent random walks and recorded the blocks they traversed and interacted with. Our results show that our model leaves \textbf{97.1\%} of the traversed blocks empty and correctly marks \textbf{95.9\%} of agent-collision blocks as walls. These findings highlight the model's ability to accurately identify key map elements, demonstrating its effectiveness in generating meaningful and navigable layouts.

\paragraph{Navigation Performance Comparison with LSTM-PPO} Next, we compare the navigation performance of our approach against LSTM-PPO\cite{wijmans2019dd}, a strong baseline in Navigation. 

We evaluate each method using 2000 navigation episodes across 100 different test maps which do not exist in training dataset. A timeout of 200 steps was imposed for each episode. We use two evaluation metrics: \textbf{Success Rate (SR)} and \textbf{Success weighted by inverse Path Length (SPL)}\cite{savva2019habitat}. Success Rate (SR) represents the percentage of episodes where the agent successfully reached the goal. And SPL defined as $\text{SPL} = S \frac{l}{\max(p, l)}$,  where \( S \) is a binary success indicator (1 if successful, 0 otherwise), \( p \) is the path length taken by the agent, and \( l \) is the optimal (shortest) path length. The agent had three possible actions at each step: move forward 0.5 meters, turn left by 30 degrees, or turn right by 30 degrees. The agent's observation space consisted of its relative position to the goal and its current orientation, mimicking a scenario where navigation relies solely on GPS and compass sensors. We also included a simple \textbf{Goal-Follower} baseline, adapted from the Habitat Challenge\cite{savva2019habitat}, where the agent directly navigates towards the goal in a straight line. The results of this comparative analysis are summarized in Table \ref{tab:navigation_performance_comparison}. Our method uses fixed parameters with a map size of 8x8, a sample number of N=10, and a CFG weight of 0.5.

For the LSTM-PPO baseline, we use a two-layer LSTM network with a hidden dimension of 512. The hyperparameters for PPO and the network architecture remain consistent with those used in \cite{wijmans2019dd}. We also use the same reward design, which includes a terminal reward and a shaped reward. The agent receives a terminal reward of \( r_T = 2.5 \) SPL if it is within 1m of the target. The shaped reward at each step is \( r_t(a_t, s_t) = -\Delta_{\text{geo dist}} - 0.01 \), where \( \Delta_{\text{geo dist}} \) is the geodesic distance change, and \(-0.01\) is a step penalty to promote efficiency. We use the fast marching method\cite{sethian1999fast} to approximate the geodesic distance.

\begin{table}[htbp]
    \centering
    \caption{The Navigation performance comparison of our method against LSTM-PPO. The numbers in parentheses indicate the environment interaction steps for LSTM-PPO and the offline dataset size for our method.}
        \begin{tabular}{lcc}
        \toprule
        Method & Success Rate (SR) & SPL \\
        \midrule
        Goal-Follower & 10.9\% & 10.1\% \\
        LSTM-PPO(100M) & 35.9\% & 28.6\% \\
        LSTM-PPO(500M) & 45.5\% & 35.5\% \\
        Our Method(0.1M) & 39.4\% & 31.4\% \\
        Our Method(1M) & 45.8\% & 36.1\% \\
        Our Method(10M) & \info{47.3\%} & \info{37.0\%} \\
        \bottomrule
    \end{tabular}
\label{tab:navigation_performance_comparison} 
\end{table}

The Goal-Follower baseline, which follows a straight-line path, performs poorly, highlighting the difficulty of our navigation tasks. In contrast, our method outperforms the best LSTM-PPO agent, which achieves its best performance after 500M interactions with the environment. We report LSTM-PPO's performance at 500M because, beyond this point, its performance experiences instability and diminishing returns. On the other hand, our model reaches the same performance while being trained on a dataset that is only 1/500th the size of what the best LSTM-PPO requires.

Furthermore, we observe that LSTM-PPO struggles with instability when the training task becomes too complex, limiting its ability to learn from episodes that demand long-term planning. Our diffusion-based approach, which uses an explicit belief representation, offers more robust and sample-efficient navigation. Additional visualizations can be found in the appendix.

\paragraph{Comparison with Deterministic and Generative Baselines}
To show that the diffusion model for map layout modeling is necessary, we included the following baselines: 1. \textbf{Deterministic Prediction Model}: In this approach, we replace diffusion with a deterministic prediction model. 2. \textbf{Trajectory Diffusion}: We train a 5-layer transformer on the same dataset to predict the next 10 steps based on the observation history of positions relative to the goal position. During testing, we use these predictions as waypoints for the agent to follow greedily, similar to our method.
\begin{table}[htbp]
    \centering
    \caption{The Navigation performance comparison of our method against LSTM-PPO.}
        \begin{tabular}{lcc}
        \toprule
        Method & Success Rate (SR) & SPL \\
        \midrule
        Trajectory Diffusion & 9.4\%& 6.2\% \\
        Deterministic Prediction & 20.4\% &  16.5\%\\
        Our Method & \info{47.3\%} & \info{37.0\%} \\
        \bottomrule
    \end{tabular}
\label{tab:navigation_performance_generative} 
\end{table}

The results demonstrate that our method significantly outperforms both deterministic and generative models on trajectory prediction in navigation tasks. The deterministic prediction model exhibits more stable performance than the trajectory generative approach, highlighting that learning a map model is usually easier than predicting future dynamics, and that generative models at the observation level may require carefully designed, high-quality datasets to be effective. However, deterministic methods' inability to capture multimodal distributions limits their effectiveness in complex environments.

\paragraph{Impact of Hyperparameters on Planning Performance} Finally, we examine the influence of key hyperparameters on navigation performance, focusing on two key parameters: \textbf{Inference Map Size (S)} and \textbf{Belief Space Sample Number (N)}. For each ablation, we report the best performance for CFG weights of 0.0 and 0.5, and sample sizes of N = 10 and 15 (except for the ablation of the number of samples).

\begin{table}
\centering
\caption{Navigation Performance with Different Map Sizes and Dataset Sizes}
\label{tab:navigation_performance}
\begin{tabular}{cccc}
\toprule
Map Size & Dataset Size & SPL (\%) & SR (\%) \\
\midrule
4   & 10M   & 33.6 & 42.1 \\
4   & 1M    & 33.1 & 41.1 \\
4   & 0.1M  & 30.0 & 37.9 \\
\midrule
8   & 10M   & \info{37.0} & \info{47.3} \\
8   & 1M    & 36.1 & 45.8 \\
8   & 0.1M  & 31.8 & 40.0 \\
\midrule
16  & 10M   & 34.9 & 44.5 \\
16  & 1M    & 35.2 & 44.3 \\
16  & 0.1M  & 30.8 & 38.2 \\
\midrule
32  & 10M   & 33.2 & 41.4 \\
32  & 1M    & 33.1 & 41.8 \\
32  & 0.1M  & 31.4 & 39.4 \\
\bottomrule
\end{tabular}
\end{table}

The results in Table \ref{tab:navigation_performance} highlight that a map size of 8 consistently delivers the best performance. Additionally, navigation performance improves as the dataset size increases for all window sizes. Increasing the map size beyond 8 either results in marginal performance reduction or no significant improvement. This suggests that larger inference maps introduce added complexity without yielding better contextual understanding when the data is limited. Furthermore, smaller map sizes, such as 4, cause the model to struggle with long-term planning. We observe that this often leads to paths with frequent back-and-forth movements due to limited context length. 

\begin{table}
\centering
\caption{Navigation Performance with Different Window Sizes and Sample Number }
\label{tab:navigation_performance_sample_number}
\begin{tabular}{cccc}
\toprule
Window Size & Sample Number & SPL (\%) & SR (\%) \\
\midrule
4   & 1   & 23.1 & 32.5 \\
4   & 5   & 31.7 & 40.2 \\
4   & 10   & 33.0 & 41.3 \\
4   & 15   & 33.6 & 42.1 \\
\midrule
8   & 1   & 23.1 & 33.0 \\
8   & 5   & 35.1 & 45.5 \\
8   & 10   & \info{37.0} & \info{47.3} \\
8   & 15   & 36.3 & 46.7 \\
\midrule
16  & 1   & 22.5 & 31.3 \\
16  & 5   & 33.4 & 42.9 \\
16  & 10   & 34.9 & 44.5 \\
16  & 15   & 34.6 & 43.1 \\
\bottomrule
\end{tabular}
\end{table}

From Table \ref{tab:navigation_performance_sample_number}, we observe significant performance improvements when increasing N from 1 to 5 and then from 5 to 10 across all window sizes. However, the performance gains from further increasing N to 15 are much smaller, with some cases showing a slight decrease. This suggests that with N=10 samples, the belief space is already adequately represented for effective navigation. Increasing the sample number beyond this point offers only marginal improvements in belief accuracy, which may lead to infeasible path suggestions and a decrease in navigation performance, particularly with larger window contexts.

\section{Conclusion}
In this paper, we presented BeliefDiffusion, a novel framework for point-goal navigation in partially observable environments that explicitly models multimodal belief states using diffusion models. By generating a distribution of plausible map layouts conditioned on observation history and integrating these layouts into a Model Predictive Control (MPC) planner, BeliefDiffusion achieved robust and efficient navigation in complex and unknown environments. Our empirical evaluation in synthetic 2D grid worlds demonstrated that BeliefDiffusion significantly outperforms both model-free reinforcement learning methods like LSTM-PPO and generative models relying on unimodal approximations or trajectory prediction, particularly in terms of sample efficiency and navigation performance.  While our results are promising, the computational demands of belief sampling for each planning step could be further improved in real-time applications. 
Future research directions include optimizing the computational efficiency of belief sampling and planning, and exploring the broader applicability of our BeliefDiffusion framework to other POMDP beyond point-goal navigation.


%
%
%
\bibliographystyle{ACM-Reference-Format}
\bibliography{aaai25}
\balance

%

\end{document}